\documentclass[journal]{IEEEtran}

\usepackage{times}
\usepackage{multirow}
\usepackage{graphicx}
\usepackage{amsmath}
\usepackage{diagbox}
\usepackage{slashbox}
\usepackage{rotating}
\usepackage{subfig}
\usepackage{color}

\begin{document}
\title{Part-Level Convolutional Neural Networks for Pedestrian Detection Using Saliency and Boundary Box Alignment}

%\author{Inyong Yun,~\IEEEmembership{Student Member,~IEEE,}
%        John~Doe,~\IEEEmembership{Fellow,~OSA,}
%        and~Jane~Doe,~\IEEEmembership{Life~Fellow,~IEEE}% <-this % stops a space
\author{Inyong Yun, Cheolkon Jung,~\IEEEmembership{Member,~IEEE,} Xinran Wang, Alfred O Hero,~\IEEEmembership{Fellow,~IEEE,} and Joongkyu Kim,~\IEEEmembership{Member,~IEEE}%
\thanks{The earlier version of this paper has been presented at the 42nd IEEE Conference on Acoustics, Speech, and Signal Processing (ICASSP), New Orleans, LA, USA, March 5-9, 2017~\cite{wang2017part}. This work was supported by the National Natural Science Foundation of China (No. 61872280) and the International S\&T Cooperation Program of China (No. 2014DFG12780).}%
\thanks{I. Yun and J. Kim are with the College of Information and Communication Engineering, Sungkyunkwan University, Suwon, Gyeonggi 16419, Korea e-mail: jkkim@skku.edu.}%
\thanks{A. O. Hero is with the Department of Electrical Engineering and Computer Science, University of Michigan, Ann Arbor, MI 48109-2122, USA e-mail: hero@umich.edu.}%
\thanks{X. Wang and C. Jung are with the School of Electronic Engineering, Xidian University, Xian, Shaanxi 710071, China e-mail: zhengzk@xidian.edu.cn}}
% <-this % stops a
%\thanks{123.}}% <-this % stops a space

\markboth{} %
{Shell \MakeLowercase{\textit{et al.}}: Bare Demo of IEEEtran.cls for IEEE Journals}

\maketitle

\begin{abstract}
Pedestrians in videos have a wide range of appearances such as body poses, occlusions, and complex backgrounds, and there exists the proposal shift problem in pedestrian detection that causes the loss of body parts such as head and legs. To address it, we propose part-level convolutional neural networks (CNN) for pedestrian detection using saliency and boundary box alignment in this paper. The proposed network consists of two sub-networks: detection and alignment. We use saliency in the detection sub-network to remove false positives such as lamp posts and trees. We adopt bounding box alignment on detection proposals in the alignment sub-network to address the proposal shift problem. First, we combine FCN and CAM to extract deep features for pedestrian detection. Then, we perform part-level CNN to recall the lost body parts. Experimental results on various datasets demonstrate that the proposed method remarkably improves accuracy in pedestrian detection and outperforms existing state-of-the-arts in terms of log average miss rate at false position per image (FPPI).

\end{abstract}


\ifCLASSOPTIONpeerreview
\begin{center} \bfseries EDICS Category: 3-BBND \end{center}
\fi
\IEEEpeerreviewmaketitle

\section{Introduction}
\label{sec:introduction}
\IEEEPARstart{O}{bject} detection is a classical task in computer vision which is an operation capturing target objects in images (or video) and feeding back the category and localization of the object. Latest solutions on object detection achieve high computing speed and accuracy. For example, YOLO~\cite{redmon2016yolo9000} produces very high performance in object detection: more than 40 frames per second (FPS) and 78 mean average precision (MAP) on PASCAL Visual Object Classes challenge 2007 (VOC2007). As a sub-field of object detection, pedestrian detection is often applied to video surveillance, automotive safety, and robotics applications. Pedestrian, a special instance in object detection, has a unique trait in videos. Pedestrians in videos have a wide variety of appearances such as body pose, clothing, lighting and occlusion, while the background might be changed in a limited range. The wide range of intra-class variety against relatively small background change has a negative effect on detectors. Above all, many detectors which work well on detecting common objects heavily suffer from occlusion in pedestrian detection, which leads to the decrease of the location quality represented by bounding boxes. Thus, occlusion handling is required to help the detectors recall test samples in different level of occlusions.

\begin{figure}[t]
\centering
\centerline{\includegraphics[width = 0.5\linewidth]{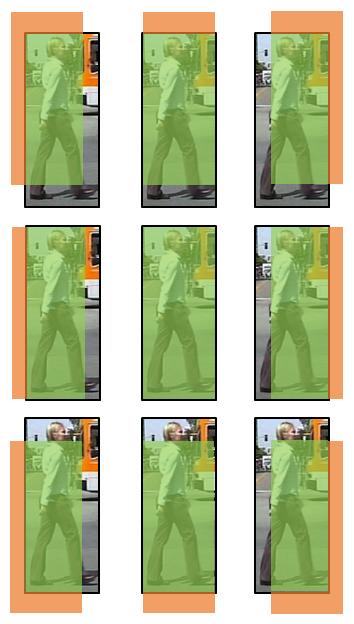}}
\caption{Proposal shift problem in pedestrian detection. The colored boxes are the detection proposals, while the black boundaries are their ground truth.}
\label{fig:shiftP}
\end{figure}
\begin{figure*}[t]
\centering
\centerline{\includegraphics[width = 0.8\linewidth]{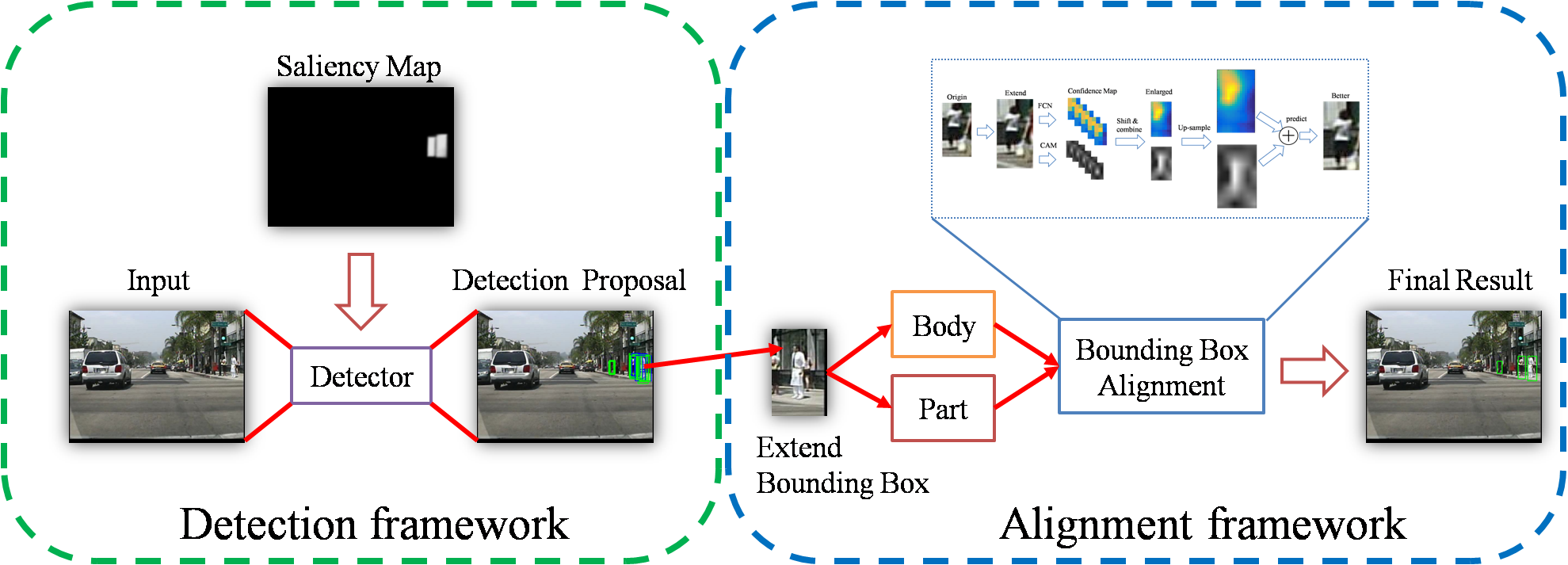}}
\caption{Whole framework of the proposed method. The proposed pedestrian network consists of two sub-networks: detection and alignment. }
\label{fig:framework}
\end{figure*}

Felzenszwalb \emph{et al.}~\cite{felzenszwalb2008a, felzenszwalb2010object} proposed a star model to search the whole image for body parts by a multi-scale sliding window technique. This work has inspired researchers to consider part detection in deep learning~\cite{ouyang2012discriminative, tian2015deep, ouyang2013joint, ouyang2013modeling, luo2014switchable}. Ouyang and Wang~\cite{ouyang2013joint} designed a unique part detection layer with 20 convolutional filters of different sizes to detect body parts of the corresponding size ratio. These deep learning-based methods assume that the detection proposals are given by conventional detectors such as SquaresChnFtrs~\cite{benenson2013seeking}. Thus, recent CNN-based pedestrian detectors~\cite{ouyang2012discriminative, tian2015deep, ouyang2013joint, ouyang2013modeling, luo2014switchable, tian2015pedestrian, angelova2015real, li2015scale, hosang2015taking} have transformed pedestrian detection to classification of the detection proposals. Thus, detectors avoid redundant exhaustive search over whole images. JointDeep~\cite{ouyang2013joint} and SDN~\cite{luo2014switchable} used "HOG+CSS" as features and a Linear SVM as a classifier to generate detection proposals (HOG: Histogram of oriented gradient, CSS: Color-self-similarity). The "HOG+CSS+SVM" proposer recalled most pedestrian candidates from images. Also, the performance of the CNN detector was improved by hard negatives generated by the "HOG+CSS+SVM" proposer. Other detection proposals were generated by ACF~\cite{dollar2014fast}, LDCF~\cite{nam2014local}, SquaresChnFtrs~\cite{benenson2013seeking}, and checkerboards~\cite{zhang2015filtered}. For the 2-stage detectors which combine detection proposal and classification are influenced significantly by the performance of detection proposers, especially for intersection over union (IoU) of bounding boxes.

In this paper, we propose part-level CNN for pedestrian detection using fully convolutional networks (FCN) and class activation map (CAM). The proposed network consists of two sub-networks of detection and alignment. In the detection sub-network, we use saliency to assign different weights to pedestrians and background. Based on saliency, we remove false positives such as lamp posts and trees from pedestrians. We adopt the alignment sub-network to recall the lost body parts caused by the detection sub-network. In the alignment sub-network, we utilize localization features of CNN such as FCN abd CAM to produce confidence maps and infer accurate accurate pedestrian location, i.e. bounding box alignment. Although FCN-based feature maps in our previous work~\cite{wang2017part} preserved the localization capability of CNN well, its output resolution was relatively low for bounding box alignment. Therefore, it was hard to obtain accurate feature maps even with upsampling used. To address the resolution problem, we add CAM into the alignment sub-network. With the help of CAM, we produce high resolution feature maps for bounding box alignment. In this work, we divide the proposed CNN detector for training into three body parts considering efficiency: head, torso and legs. In our previous work~\cite{wang2017part}, we divided it into five parts of head, left torso, right torso, left leg and right leg. Moreover, we utilize the detection sub-network to obtain pedestrian proposals, while our previous work~\cite{wang2017part} used SquaresChnFtrs~\cite{benenson2013seeking} based on a combination of conventional hand-crafted features. Experimental results show that the proposed method effectively removes false positives by saliency as well as successfully recall the lost body parts by boundary box alignment. The proposed method achieves 10\% performance improvement in pedestrian detection over our previous work~\cite{wang2017part}. Fig.~\ref{fig:framework} illustrates the whole framework of the proposed method. 

Compared with the existing methods, main contributions of this paper are as follows:

\begin{itemize}
\item[$\bullet$] We use saliency in the detection sub-network to remove background areas such as lamp posts and trees from pedestrians.
\item[$\bullet$] We combine FCN and CAM into the alignment sub-network to enhance the resolution of confidence maps and successfully recall the lost body parts.
\end{itemize}

The rest of this paper is organized as follows.
Section~\ref{sec:RelatedWorks} relevant research trends.In Section~\ref{sec:ProposedMethod}, the proposed method are described in detail. Section~\ref{sec:Experiments} experimentally compares the proposed method with existing methods. Section~\ref{sec:Conclusion} draws conclusions.

\section{Related Work}
\label{sec:RelatedWorks}
Up to the present, researchers have proposed many outstanding works for pedestrian detection, and in this section we mainly focus on deep learning models. The first deep model was an unsupervised deep model proposed by Sermanet \emph{et al.}~\cite{sermanet2013pedestrian} to consider limited training data. This model used a few tricks: 1) Multi-stage features, 2) connections to skip layers and integrate global shape information with local distinctive motif information, 3) unsupervised method based on convolutional sparse coding to pre-train the filters at each stage. A series of methods~\cite{ouyang2012discriminative, ouyang2013joint, ouyang2013modeling, luo2014switchable} combined part detection and deep models to improve the detection accuracy in body part occlusion. DBN-Isol~\cite{ouyang2012discriminative} proposed the deformable part model (DPM)~\cite{felzenszwalb2010object} based on a deep belief network to estimate the visibility of pedestrians. JointDeep~\cite{ouyang2013joint} was a deep model that was composed of feature extraction, occlusion handling, deformation and classification in a single network. MultiSDP~\cite{zeng2013multi} built a multi-stage classifier to deal with complex distributed samples in pedestrian datasets. SDN~\cite{luo2014switchable} used switchable Restricted Boltzmann Machines (RBMs) to extract high-level features for body parts. They divided human body into three parts: head-shoulder, upper-body and lower-body.
Tian \emph{et al.}~\cite{tian2015pedestrian} introduced datasets for scene labeling which contained city street scenes to aid the detector for distinguishing background from the proposals. The idea was that the scene labeling datasets contained information similar to the background in pedestrian datasets. Considering part detection, Tian \emph{et al.}~\cite{tian2015deep} also proposed DeepParts to handle occlusion with an extensive body part pool. In this method, SVM detector was not used directly for the CNN output due to its small improvement. Moreover, general object detectors~\cite{girshick2014rich} have been applied to pedestrian detection. Hosang \emph{et al.}~\cite{hosang2015taking} analyzed the feasibility of the region-based CNN~\cite{girshick2014rich} (R-CNN) framework for the pedestrian detection task. They adopted SquaresChnFtrs~\cite{benenson2013seeking}, i.e. a stand-alone pedestrian detector, as the detection proposer and a R-CNN model for classification. Following R-CNN, region proposal network (RPN) built in faster R-CNN~\cite{ren2015faster} produced detection proposals by the network itself.

\section{Proposed Method}
\label{sec:ProposedMethod}
The proposed pedestrian detection framework consists of two sub-networks: detection and alignment. We use a proposal-and-classification approach to detect pedestrians with multi-scales. To get detection prosals, we perform fast pedestrian detection in the detection sub-network based on region proposal network (RPN). To remove false positives, we use saliency in the detection sub-network. Then, we align bounding boxes in the alignment sub-network to recall the lost body parts caused by the detection sub-network. We combine FCN and CAM into the alignment sub-network for accurate pedestrian localization.

\subsection{Detection Framework}
\label{subsec:DetectionFramework}

\begin{figure*}[t]
\centering
\centerline{\includegraphics[width = 0.7\linewidth]{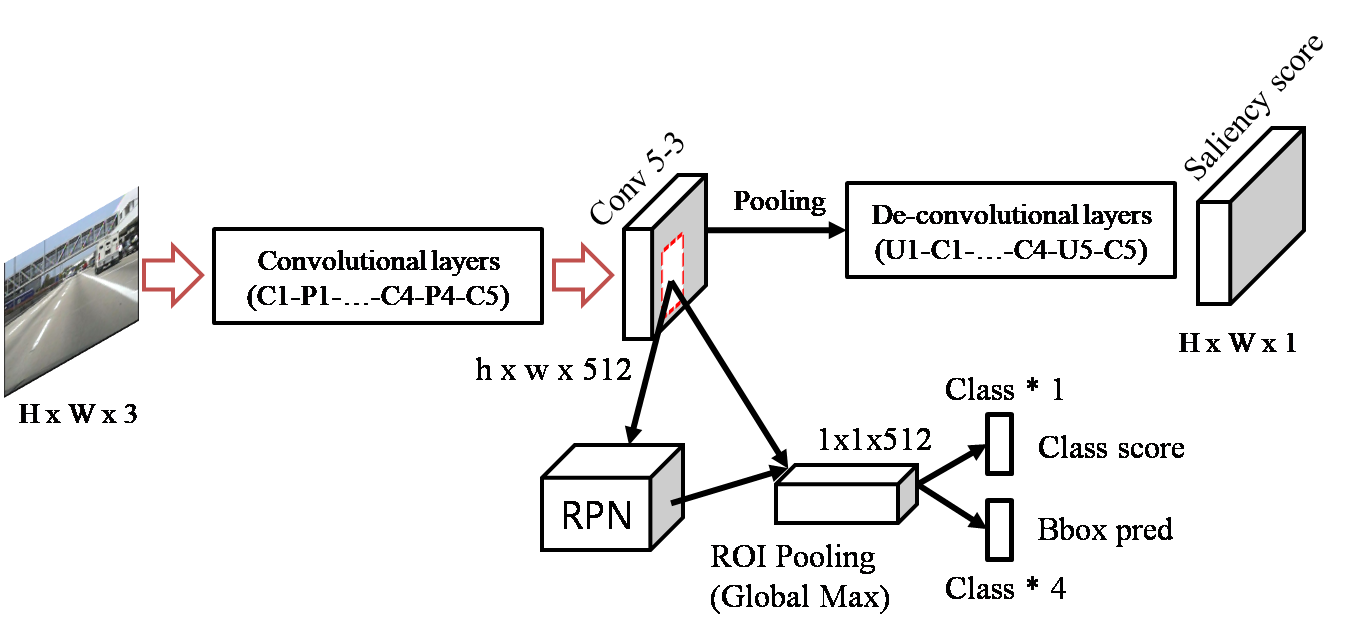}}
\caption{Architecture of the proposed detection sub-network. }
\label{fig:detection}
\end{figure*}

 \textbf{Network Architecture:} The first stage is to generate detection proposals. As shown in Fig.~\ref{fig:detection}, the detection sub-network consists of five convolutional units, one fully-connected (FC) layer, and one global max pooling (GMP) layer for classification and localization. The five convolutional units are configured similar to the VGG-16 network~\cite{simonyan2014very}. Each convolutional unit consists of two or three $3 \times 3$ convolutional layers and one max-pooling layer. The fifth convolutional unit is connected by a global max pooling layer instead of a max pooling layer. These convolutional layer produces a feature map of size $1 \times 1 \times 512$. The feature map is connected to the FC layer, which is separated by two output layers. The first output layer is the classification layer, while the second output layer is the bounding box regression layer. This output layer architecture is taken from Faster R-CNN~\cite{ren2015faster}. For the network training, the loss ($L_{d}$) is defined as follows:

\begin{equation}
	L_{d} = L_{d}^{cls} + L_{d}^{bbox}
\end{equation}

where $L_{d}^{cls}$ is the classification loss, i.e. softmax-log loss, and $L_{d}^{bbox}$ is the bounding box regression loss, i.e. smooth L1 loss.

Also, we add three convolutional layers and five deconvolutional blocks in the saliency network since the last pooling layer in the detection sub-network to get saliency maps for pedestrians. The deconvolutional block consists of one bilinear upsampling layer, one or three convolutional units. The layer configuration of the deconvolution block for the saliency network is described in Table~\ref{table:deconv_block_details}. In the last deconvolution block, the output value is limited to 0 to 1 using sigmoid function. For the network training, we calculate the saliency loss $L_{s}$ by simple Euclidean distance from the ground truth.

\begin{table}[t]
\caption{Layer configuration of the deconvolutional block for the saliency network. Input size: $600 \times 800$. Change of in/out channels: $\to$. Change of layer size: $\downarrow, \uparrow$. Data flow: $\leftarrow$.}
\small\addtolength{\tabcolsep}{-3pt}
\begin{center}
\begin{tabular}[c]{|c|c|c|c|c|}
\hline
\bf{Layer} & \bf{Filter} & \bf{Size ($w\times h$)} & \bf{Output} & \bf{etc.}  \\
\hline
$pool$ 5    & 2 $\times$ 2 &  $18 \times 25 $ & 512 & $\leftarrow conv$ 5-3, $\downarrow$\\
$conv$ 6-1  & 3 $\times$ 3 &  $18 \times 25$ & 512$\to$1024 & \\
$conv$ 6-2  & 3 $\times$ 3 &  $18 \times 25$ & 1024 & \\
$conv$ 6-3  & 3 $\times$ 3 &  $18 \times 25$ & 1024 & \\
\hline
$upsample$ 1  & -          & $35 \times 49$  & 1024 & size $\uparrow$ \\
$conv$ 7-1  & 3 $\times$ 3 &  $35 \times 49$ & 1024$\to$512 & \\
$conv$ 7-2  & 3 $\times$ 3 &  $35 \times 49$ & 512 & \\
\hline
$upsample$ 2  & - & $69 \times 97$ & 512 & size $\uparrow$ \\
$conv$ 8-1  & 3 $\times$ 3 &  $69 \times 97$ & 512 $\to$ 256 & \\
$conv$ 8-2  & 3 $\times$ 3 &  $69 \times 97$ & 256 & \\
\hline
$upsample$ 3  & -          & $137 \times 193$  & 256 & size $\uparrow$ \\
$conv$ 9-1  & 3 $\times$ 3 &  $137 \times 193$ & 256$\to$128 & \\
\hline
$upsample$ 4  & -          & $273 \times 385$ & 128 & size $\uparrow$ \\
$conv$ 10-1  & 3 $\times$ 3 &  $273 \times 385$ & 128$\to$64 & \\
\hline
$upsample$ 5  & -          & $ 545\times 769$ & 64 & size $\uparrow$ \\
$conv$ 11-1  & 3 $\times$ 3 &  $545 \times 769$ & 64 $\to$ 32 & \\
$conv$ 11-2  & 3 $\times$ 3 &  $545 \times 769$ & 32 $\to$ 1 & \\
\hline
\end{tabular}
\end{center}
\label{table:deconv_block_details}
\end{table}

For detection proposals, we train the detection sub-network jointly with the saliency network by optimizing the following combined loss function:

\begin{equation}
L = L_{d} + L_{s}
\end{equation}

where $L_{d}$ and $L_{s}$ are losses of the detection network and of the saliency network, respectively. \\

\textbf{Detection Proposal:} We use Faster R-CNN~\cite{ren2015faster} to extract detection proposals for pedestrians. However, the detection results include some false positives such as vehicle parts, trees, and post lamps. To remove them, we apply different weights to the background and foreground so that the detector focuses on the pedestrian area. To determine the weight, we obtain pedestrian saliency maps using the saliency network from the input image. We update the class probability (score) using saliency map as follows:

\begin{equation}
	f_{w}(b) = f(b)* w_{f}
\end{equation}

The weight $w_{f}$ is defined as follows:

\begin{equation}
w_{f} = \lbrace
\begin{array}{cl}
1 & if \quad  f(b) > th_{b} \\
\frac{1}{N}\sum_{x,y \in b}s(x,y) & otherwise,
\end{array}
\end{equation}

\begin{figure}[t]
\centering
\subfloat[]{
\label{Fig.sub.1}
\includegraphics[width = 0.15\textwidth]{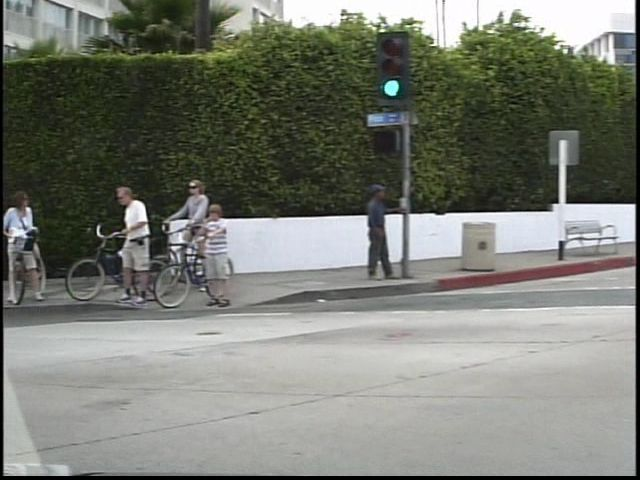}}
\subfloat[]{
\label{Fig.sub.1}
\includegraphics[width = 0.15\textwidth]{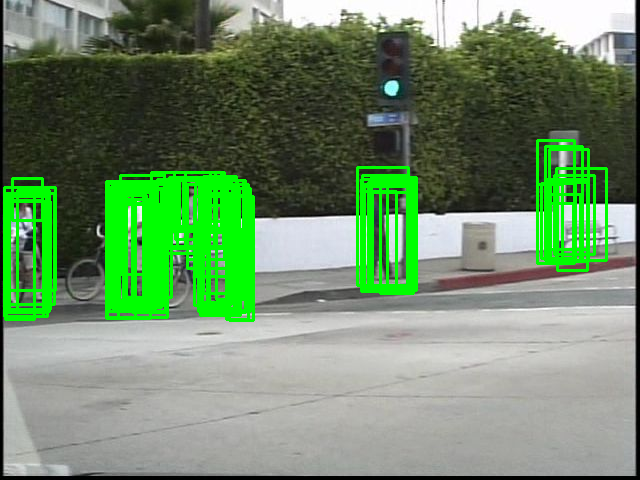}}
\subfloat[]{
\label{Fig.sub.1}
\includegraphics[width = 0.15\textwidth]{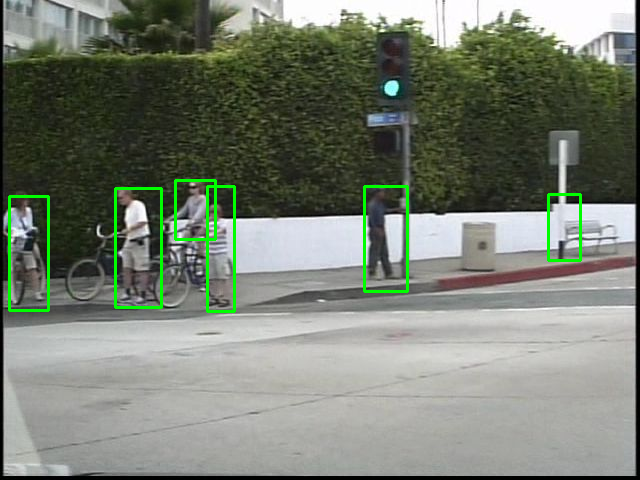}}

\subfloat[]{
\label{Fig.sub.1}
\includegraphics[width = 0.15\textwidth]{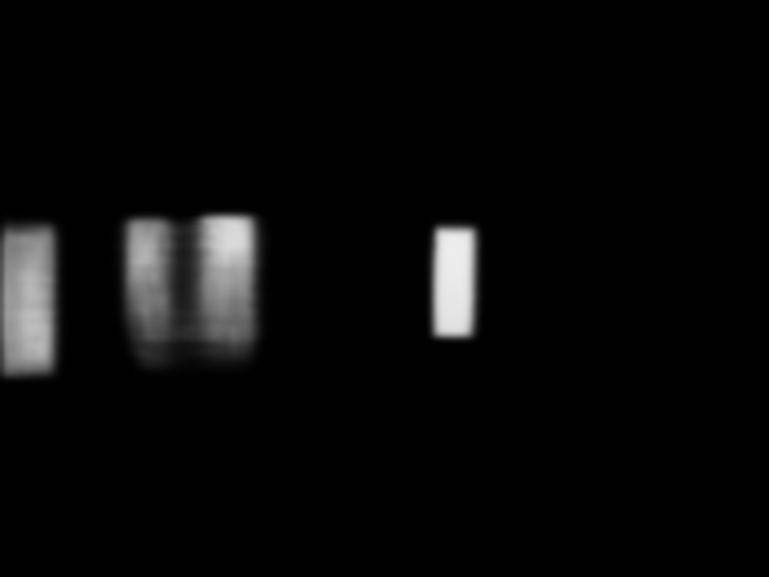}}
\subfloat[]{
\label{Fig.sub.1}
\includegraphics[width = 0.15\textwidth]{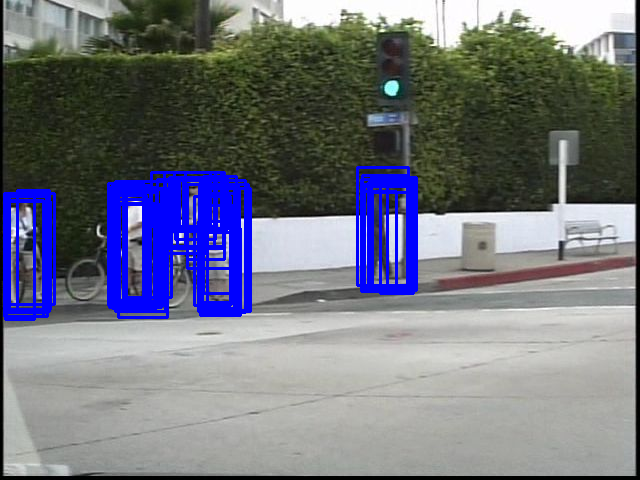}}
\subfloat[]{
\label{Fig.sub.1}
\includegraphics[width = 0.15\textwidth]{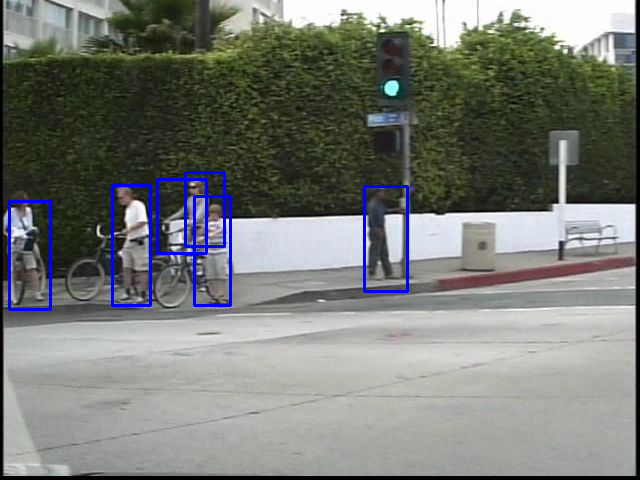}}
\caption{Examples of detection proposal with saliency weight. (a) Input image, (b) detection proposal (w/o $w_{f}$), (c) NMS result of (b), (d) Saliency map, (e) detection proposal (with $w_{f}$), (f) NMS result of (e) }
\label{fig:withSal}
\end{figure}

where $b$ is bounding boxes of proposals, $s(x,y)$ is saliency scores in the position ($x$, $y$), and $f(b)$ is class scores of the selected bounding box. $th_{b}$ is the threshold value for distinguishing between foreground and background. The new class score $f_{w}(b)$ is calculated by the product of the weight value $w_{f}$ and the bounding box score $f(b)$. Then, we use non-max suppression (NMS)~\cite{ren2015faster} to determine the final detection proposal samples. Fig.~\ref{fig:withSal} shows some examples of the detection proposal samples generated by the proposed method.

\subsection{Alignment Framework}
\label{subsec:AlignmentFramework}
\begin{figure*}[t]
\centering
\centerline{\includegraphics[width = 0.7\linewidth]{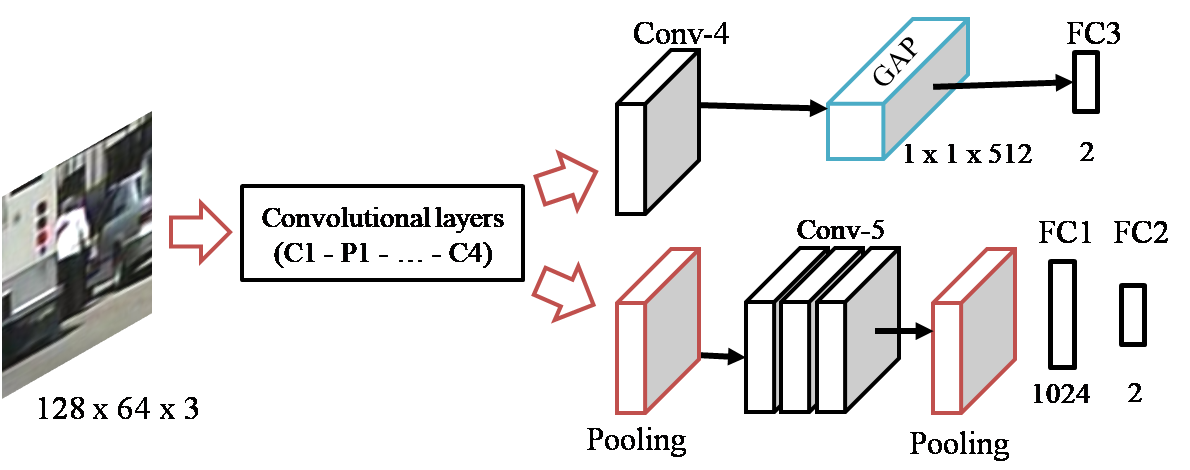}}
\caption{Network architecture of the proposed part-level detector based on VGG-16 network with class activation map}
\label{fig:partD}
\end{figure*}

\textbf{Network Architecture:} The second stage is to align the bounding box using part-level detector. Our part-level detector is a combination of one root detector which detects root position of pedestrians and three part-level detectors which detect human body parts of head, torso, and legs. The root/part detector networks are configured similar to VGG-16 network. As shown in Fig~\ref{fig:partD}, the alignment sub-network has two output layers: One is the output layer to obtain FCN and the other is the output layer to obtain CAM with global average pooling.

Our root-detector produces confidence score and root position for detection proposals. Bounding box alignment is performed on the root detector, and we treat this updated position of the aligned bounding box as an anchor position, i.e. the final position. Similarly, part confidence score and part position are produced by each part-level detector. Note that the part detection stage is implemented based on the updated position. Theoretically, bounding box alignment helps the proposed detector by better detection proposals as well as recall the lost body parts which is out of the ground truth. We compute a weighted sum of the confidence scores with a spatial distance penalty term as the final confidence score of a detection proposal.  \\

\textbf{Converting CNN into FCN/CAM:} In general, detectors suffer from low detection IoU such as R-CNN, which causes poor localization quality of the detection proposals. In this work, we name it as the proposal shift problem. Hosang \emph{et al.}~\cite{hosang2015taking} reported that the best detection proposal method SpatialPooling+~\cite{paisitkriangkrai2016pedestrian} recalled 93\% samples with 0.5 IoU threshold while only recalling 10\% samples with 0.9 IoU threshold. Zhang \emph{et al.}~\cite{Shanshan2016CVPR} clustered all false positives in 3 categories, and localization quality is one of the main source of false positives. Detection proposals shift the position of samples by various direction and distance. As shown in Fig.~\ref{fig:shiftP}, body parts frequently appear out of the region of the detection proposal, which leads to bad detection response: low confidence score and/or IoU. Thus, we introduce a novel technique based on FCN and CAM to align the bounding boxes. According to the response of FCN and CAM, we generate much larger heat maps. Then, we predict the new position of pedestrians.

To perform bounding box alignment, a larger detection region is needed as the input of the detector. In this larger detection region, our root detector outputs a coarse position of a pedestrian. We simply convert root/part networks into FCN version and generate root/part CAM to get coarse position information, named as root/part-net. In root/part-net, the last pooling layer is fully  connected with \verb"FC1" by an inner product weight matrix. Thus, the size of the input image is supposed to be fixed. With the trained root/part-net, we change the shape and dimension of the parameters between the last pooling layer and \verb"FC1" to make these weight matrix convolute on the large feature map. By expanding 25\% from the size of bounding box and changing the size of the input image to $160 \times 96$, we obtain a confidence score heat map ($C_{fcn}$) of the size  $5 \times 3$.  According to the study on visualizing deep learning~\cite{ZeilerF13Vis1, MahendranV14Vis2}, the deeper the layers, the more abstract the information extracted. That is, the object neurons respond to transform simple edges to advanced information. We use the advanced information to identify categories in input images~\cite{zhou2016learning}. As shown in Fig.~\ref{fig:partD}, the global average pooling (GAP) produces the average space value of the attribute map of each unit in the 4th convolutional layer, and uses the weighted sum of the attribute values to output the final object position. The weighted sum of confidence class activation map ($C_{cam}$) is as follows:

\begin{equation}
 C_{cam} = \sum_{x,y}\sum_{k}w_{k}^{c}f_{k}(x,y)
\end{equation}

where $f_{k}(x,y)$ denotes the activation of the unit $k$ in the 4th convolutional layer for the input images, and $w_{k}^{c}$ is the weighted value corresponding to the class position in the unit $k$. Based on the previous research~\cite{zhou2016learning}, it is expected that each unit in the convolutional layer is activated by the visual pattern within the receptive field. \\

 \textbf{Shift-and-Stitch for a Larger Confidence Map:} To predict a coarse position of a pedestrian in the large detection region, a higher resolution of $C_{fcn}$ and $C_{cam}$ are needed. We use a simple trick to obtain it as follows. Since there are total $s=32$ pixels between every step, we shift the proposal windows by $f$ steps on the horizontal and vertical axis uniformly and make total distance no more than 32 pixels. This means that the shift distance of every stride is $s/f$. Also, we take root-FCN as an example, and root-FCN generates a $5 \times 3$ heat map by every step interlacing all $f^{2}$ outputs according to the relative direction of every shift-and-stitch. As a result, a $(5\cdot{f})\times(3\cdot{f})$ heat map is generated.

 \begin{figure*}[t]
\centering
\centerline{\includegraphics[width = 0.8\linewidth]{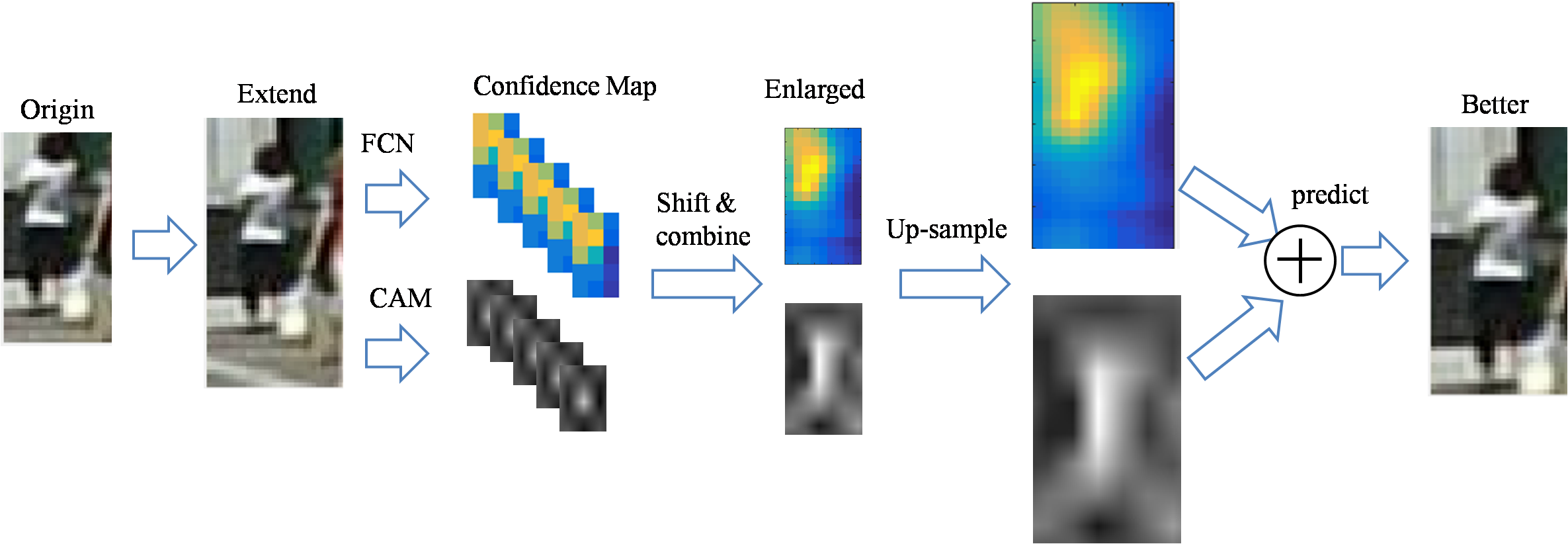}}
\caption{Pipeline for bounding box alignment. Origin: Original bounding box. The pedestrian is localized at the top left corner of a bounding box. Extend: Enlarged bounding box. Confidence map: Output of FCN and CAM. Better: Aligned bounding box. The lost head part is recalled and thus the pedestrian is accurately localized. }
\label{fig:example}
\end{figure*}

Once got a larger $C_{fcn}$ and $C_{cam}$, we apply a simple up-sampling method to produce a nice aspect ratio score heat map which equals to the aspect ratio of the input region. In this way, shift direction for the target position is calculated without a stretch operation. A coarse body position is estimated by selecting a region having the largest average value in the up-sampled $C_{fcn}$ and $C_{cam}$. We use an enlarging ratio parameter $L$ to determine the size of the target bounding box. Width/height of the rectangle $w/h$ is obtained by multiplying $L$ with the width/height of the input region $W$/$H$.

\begin{equation}
w/h=L\cdot W/H
\end{equation}

Define the coarse position in the input large region as $(x_p, y_p)$, the original position as $(x_o, y_o)$.
Then, we update $x$ by

\begin{equation}
\Delta{x}_{fcn}=\frac{2\times\sum_{i=1}^{n} (C_{fcn,i}^{t}-C_{fcn,i}^{o})^{2}}{\sum_{i=1}^n {C_{fcn,i}^{t}}^{2}+\sum_{i=1}^{n}{C_{fcn,i}^{o}}^{2}}\ast (x_{p}-x_{o})
\end{equation}

where $C_{fcn,i}^{t}$ is the value of the $i$-th element in the target rectangle in the confidence score heat map, $C_{fcn,i}^{o}$ is the value of the $i$-th element in the original rectangle, and $n$ is the total number of elements in the rectangles. $\Delta{x}_{cam}, \Delta{y}_{fcn},\Delta{y}_{cam} $ is obtained in the same way. The position of the detection proposal is updated by

\begin{equation}
x_a=x_o+\frac{\Delta{x}_{fcn}+\Delta{x}_{cam}}{2}
\end{equation}

$y_a$ is also updated in the same way. The updated position of the detection proposal $(x_a, y_a)$ is named as anchor position. Based on the anchor position $(x_a, y_a)$, our part-level detector is operated to yield part scores and part positions. \\

\textbf{Part Merging:} Part detection is considered in the alignment sub-network. The part detector has a different receptive size filter for the aligned BB generated by the root detector. Part score $score_p$ and part position $(x_p, y_p)$ that indicate the possibility and area the part appearance, respectively, are produced by each of the part detectors. The final detection score is defined as:

\begin{equation}
score = score_{root}+\sum_{i=\{parts\}}{w_i}\ast(score_i + P_i)
\end{equation}

where $score_{root}$ is the output score of the body detector; $score_i$ is the output score of three body parts; $w_i$ is the weight that indicates the importance of part scores, and we set $\sum_{i=\{parts\}}w_{i}=1$ in this work. $P_i$ is the penalty term of the spatial distance between anchor position and part position:

\begin{equation}
P=a\ast(|x_{p}-x_{a}|+|y_{p}-y_{a}|)+b\ast(|x_{p}-x_{a}|^{2}-|y_{p}-y_{a}|^{2})
\end{equation}

where $a$ and $b$ are weights of the penalty term that balance the orientation and geometrical shifting distance; $(x_a, y_a)$ is the anchor position which is the position of an aligned detection proposal. For position of the detection, we simply use the anchor position as the final position.

\subsection{Implementation Details}
\label{subsec:Implementation}
\textbf{Target Labels for Training data:} Currently, the datasets such as Caltech~\cite{Dollar2012PAMI}, INRIA~\cite{dalal2005histograms} and ETH~\cite{eth_biwi_00534} do not provide part-level and saliency annotations. Inspired by~\cite{felzenszwalb2008a,felzenszwalb2010object}, we have cropped all ground truth into three parts uniformly and assign their corresponding part labels automatically to generate training data for our part detectors. We have trained part detectors for three body parts of head, torso and legs. In Caltech pedestrian dataset, every frame in which a given sample is visible has two bounding boxes. One bounding box indicates the full extent of the entire body (BB-full), while the other is for visible region (BB-vis). For part detectors, we only select BB-vis for part division to avoid collecting background regions into positives. To generate training data for saliency, we draw a white rectangles in the black background using ground truth bounding boxes. \\

\textbf{Initialization and Settings for Training:} We have implemented the entire learning network using TensorFlow~\cite{tensorflow16tensorflow}. We have performed the learning of the proposed network on a PC with NVIDIA GTX 1080ti of 11GB memory. We have initialized the parameters of convolutional units from VGG-16~\cite{simonyan2014very}, which is pre-trained on ImageNet dataset. If not belong to VGG-16 network, Xivier initialization method~\cite{Xavier10init} is used for the weight initialization of the proposed network. For optimization, we have used ADAM optimizer~\cite{KingmaB14Adam} for learning with the learning rate 0.001 and the iteration epoch 15. Also, we avoid overfitting, and apply a dropout technique~\cite{srivastava14dropout} to the final fully-connected layer with the probability 0.5 for normalization.

\section{Experimental Results}
\label{sec:Experiments}

\subsection{Datasets and Benchmark}
\label{subsec:Dataset}

\begin{figure}[t]
 \centering
\begin{minipage}[b]{0.32\linewidth}
  \centering
  \centerline{\includegraphics[width = 1.0\textwidth]{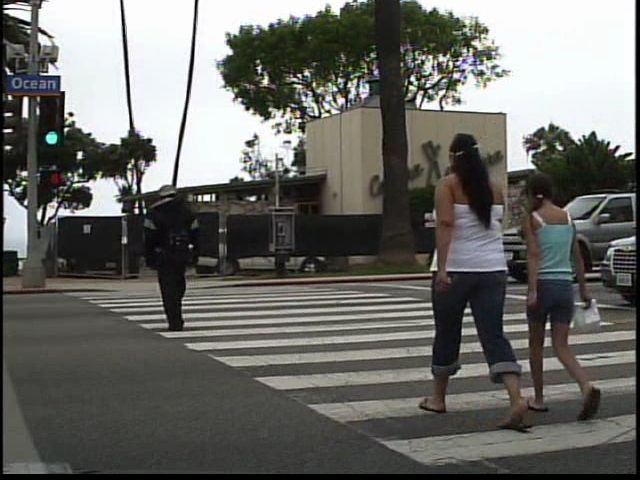}}
  \centerline{(a) }
\end{minipage}
\begin{minipage}[b]{0.32\linewidth}
  \centering
  \centerline{\includegraphics[width = 1.0\textwidth]{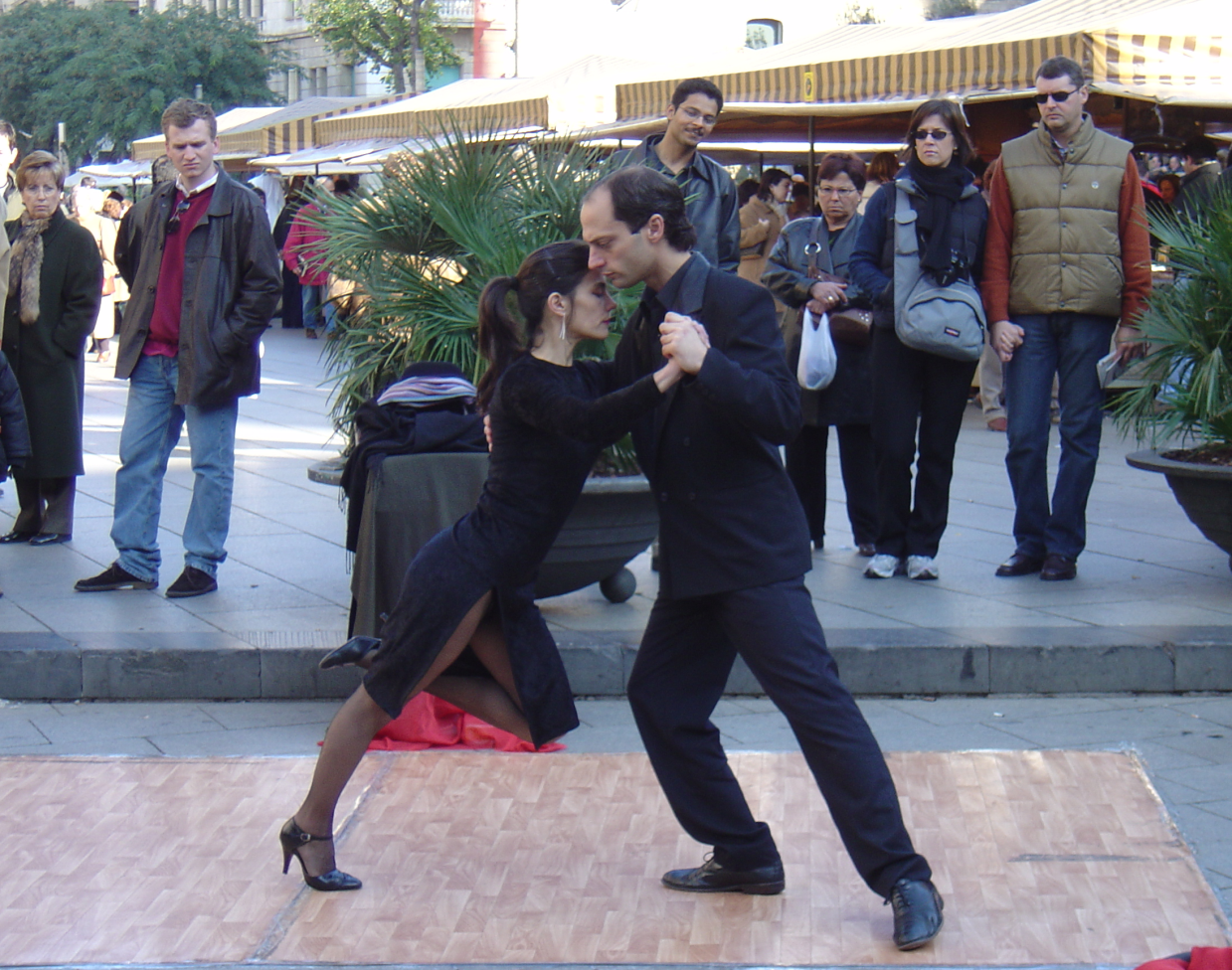}}
  \centerline{(b) }
\end{minipage}
\begin{minipage}[b]{0.32\linewidth}
  \centering
  \centerline{\includegraphics[width = 1.0\textwidth]{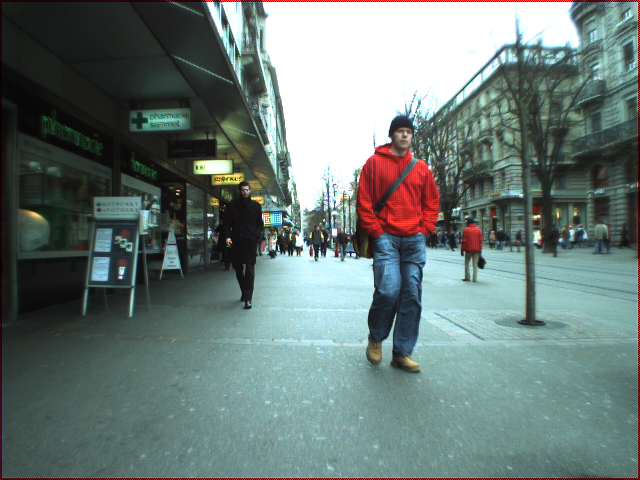}}
  \centerline{(c) }
\end{minipage}

\caption{Three datasets for experiments. (a) Caltech-USA. (b) INRIA. (c) ETH. }
\label{fig:dataset}
\end{figure}

As shown in Fig.~\ref{fig:dataset}, we evaluate performance of the proposed method on three datsets: Caltech~\cite{Dollar2012PAMI}, INRIA~\cite{dalal2005histograms} and ETH~\cite{eth_biwi_00534}.

\textbf{Caltech-USA:} This dataset~\cite{Dollar2012PAMI} consists of approximately 10 hours of $640 \times 480$ 30Hz video taken from a vehicle driving through regular traffic in an urban environment. About 250,000 frames (in 137 approximately minute long segments) with a total 350,000 bounding boxes and 2,300 unique pedestrians have been annotated. We use every 3rd frame to extract training data followed by~\cite{hosang2015taking} and~\cite{nam2014local}. The 4,024 standard testing dataset (sampling every 30th frame from test videos) are evaluated.   \\

\textbf{INRIA:} This dataset~\cite{dalal2005histograms} consists of 1,382 training images and 288 testing images taken from a personal digital image collections or the web using Google images. Only upright person (with person height $> 100$ pixels ) have been annotated. The original positive images are of very high resolution (approximately $2592 \times 1944$ pixels), and thus we have cropped these images to highlight persons. Our model is trained with all training images and evaluated on the 288 testing images.   \\

\textbf{ETH:} This dataset~\cite{eth_biwi_00534} consists of 1,450 training images and 354 testing images with a resolution of $640 \times 480$ (bayered). The dataset provides the camera calibration and annotations of pedestrian bounding boxes. \\

To evaluate the proposed pedestrian detection method, we mainly use a reasonable subset~\cite{Dollar2012PAMI, li2018scale} which contains pedestrians that have over 50 pixels height and over 65\% visibility. We perform evaluations on the final output: List of detected bounding boxes with category scores. We use the standard parameter setting on Caltech dataset. We use log-average miss rate to evaluate the detector's performance computed by average miss rate at false positive per image (FPPI) rates evenly spaced in log-space in the range $10^{-2}$ to $10^{0}$. The area that overlap with the ground truth exceeds 50\% is set to the true as follows:

\begin{equation}
 overlap = \frac{area(BB_{dt} \bigcap BB_{gt})}{area(BB_{dt} \bigcup BB_{gt})} > 0.5
\end{equation}

where $BB_{dt}$ and $BB_{gt}$ are detection bounding box and ground truth bounding box, respectively.

\begin{table}[t]
\caption{Performance evaluation on Caltech dataset (Unit=\%). Proposed I: "Detection Proposal + Saliency". Proposed II: "Proposed I + Shift Handling + Part Detectors". }
%\small\addtolength{\tabcolsep}{-3pt}
\begin{center}
\begin{tabular}[c]{|c|c|c|c|}
\hline
\bf{Subset} & \cite{wang2017part} & Proposed I & Proposed II  \\
\hline
Reasonable & 22.52 & 18.82 & 12.40 \\
\hline
Scale=Large & 8.87 & 8.70 & 4.50 \\
\hline
Scale=Near & 11.96 & 10.98 & 6.03 \\
\hline
Scale=Medium & 65.54 & 53.71 & 53.71 \\
\hline
Occ=None & 19.69 & 16.03 & 11.43 \\
\hline
Occ=Partial & 43.74 & 36.32 & 16.68 \\
\hline
\end{tabular}
\end{center}
\label{table:result1}
\end{table}

\subsection{Performance of Part-Level Detectors}
\label{subsec:per-part}

We conduct a set of experiments on Caltech dataset to investigate the detection accuracy of the proposed method. We provide the performance of the pedestrian detection on saliency weights, shift handling, and part merging. When saliency weights are applied to the detection proposals, FPPI is 18.82\% ('Proposed I' in Table~\ref{table:result1}). In comparison with the previous results, the saliency weights help to ensure the correct detection proposal as shown in Fig.~\ref{fig:withSal}). We also confirm that FPPI decreases 12.40\% by solving the proposal shift problem when the bounding box alignment is applied ('Proposed II' in Table~\ref{table:result1}). We apply part-level detection to the larger detection region. Part-level detectors are able to recall the lost body parts beyond detection proposals. With the aligned anchor positions, part positions are more accurate by localizing the largest area with average scores. The spatial distance penalty term between anchor and part positions is very effective to consider the proposal shift problem.

 \begin{figure}[t]
\centering
\subfloat[]{
\label{Fig.sub.10}
\includegraphics[width = 0.11\textwidth]{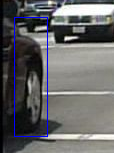}
\includegraphics[width = 0.11\textwidth]{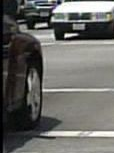}} \quad
\subfloat[]{
\label{Fig.sub.11}
\includegraphics[width = 0.11\textwidth]{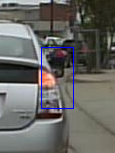}
\includegraphics[width = 0.11\textwidth]{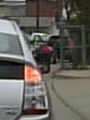}}\\
\subfloat[]{
\label{Fig.sub.12}
\includegraphics[width = 0.11\textwidth]{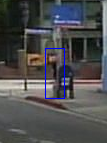}
\includegraphics[width = 0.11\textwidth]{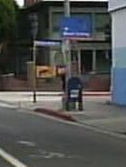}}\quad
\subfloat[]{
\label{Fig.sub.13}
\includegraphics[width = 0.11\textwidth]{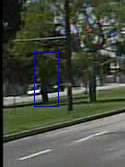}
\includegraphics[width = 0.11\textwidth]{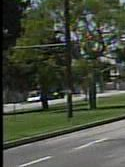}}
\caption{Some successful detection results. The left and right images show the detection results of 'basic (without saliency)' and 'proposed (with saliency)', respectively. Blue box: False positive. Best viewed in color.}
\label{fig:result3}
\end{figure}

\begin{figure}[t]
\centering
%\begin{minipage}{1.0\linewidth}
%\subfloat[]{
%\label{Fig.sub.1}
%\includegraphics[width = 0.15\textwidth]{figures/result_1_1.png}
%\includegraphics[width = 0.15\textwidth]{figures/result_1_2.png}}
\subfloat[]{
\label{Fig.sub.8}
\includegraphics[width = 0.11\textwidth]{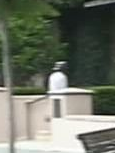}
\includegraphics[width = 0.11\textwidth]{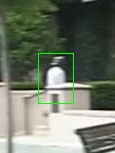}} \quad
\subfloat[]{
\label{Fig.sub.9}
\includegraphics[width = 0.11\textwidth]{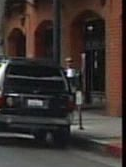}
\includegraphics[width = 0.11\textwidth]{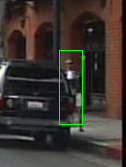}}
%\end{minipage}
\caption{Some successful detection results. The left and right images show the detection results of Basic (without saliency) and Proposed (with "Saliency + Shift Handling + Part Detectors"). Green box: True positive. Best viewed in color.}
\label{fig:result2}
\end{figure}

\begin{figure*}[t]
\centering
%\begin{minipage}{0.1\linewidth}
%\centering
\subfloat[]{
\label{Fig.sub.1}
\includegraphics[width = 0.12\textwidth]{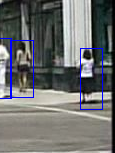}
\includegraphics[width = 0.12\textwidth]{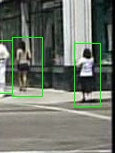}} %\quad
\subfloat[]{
\label{Fig.sub.2}
\includegraphics[width = 0.12\textwidth]{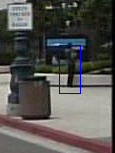}
\includegraphics[width = 0.12\textwidth]{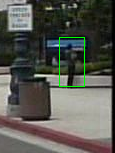}}%\quad
\subfloat[]{
\label{Fig.sub.3}
\includegraphics[width = 0.12\textwidth]{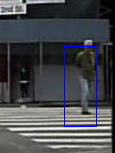}
\includegraphics[width = 0.12\textwidth]{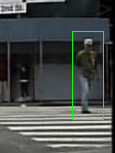}}%\quad
\subfloat[]{
\label{Fig.sub.4}
\includegraphics[width = 0.12\textwidth]{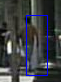}
\includegraphics[width = 0.12\textwidth]{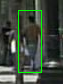}}\\
\subfloat[]{
\label{Fig.sub.5}
\includegraphics[width = 0.12\textwidth]{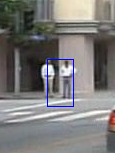}
\includegraphics[width = 0.12\textwidth]{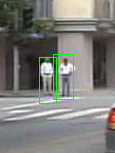}}\quad
\subfloat[]{
\label{Fig.sub.6}
\includegraphics[width = 0.17\textwidth]{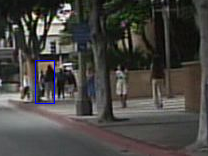}
\includegraphics[width = 0.17\textwidth]{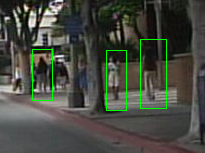}}\quad
\subfloat[]{
\label{Fig.sub.7}
\includegraphics[width = 0.17\textwidth]{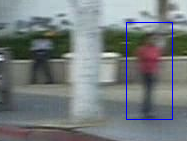}
\includegraphics[width = 0.17\textwidth]{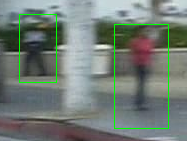}}
\caption{Successful detection results by the proposed method. The left and right images show the detection results of Basic (without saliency) and Proposed (with "Saliency + Shift Handling"), respectively. Blue box: Basic detection result. Green box: Proposed detection result.}
\label{fig:result1}
\end{figure*}

We provide some successful detection results by adding saliency (Figs.~\ref{fig:result3} and ~\ref{fig:result1}), shift handling (Fig.~\ref{fig:result1}), and part-level detector (Fig.~\ref{fig:result2}). The saliency helps to distinguish background components similar to pedestrians. Without saliency, it is easy to falsely detect car parts (Figs.~\ref{Fig.sub.10} and \ref{Fig.sub.11}) or trees (Figs.~\ref{Fig.sub.12} and \ref{Fig.sub.13}) as pedestrians because cars or trees have similar shapes to pedestrians. The proposed method improves the detection performance by separating one box with two pedestrians (Fig.~\ref{Fig.sub.5}) and detecting pedestrians blurred by motion (Fig.~\ref{Fig.sub.7}). Moreover, the proposed method recalls the lost body parts by bounding box alignment as shown in Figs.~\ref{Fig.sub.1}-\ref{Fig.sub.4}). The part-level detector is able to detect partially-occluded or low-resolution pedestrians that the upper body is visible (Fig.~\ref{Fig.sub.8}) and the body parts are occluded (Fig.~\ref{Fig.sub.9}).

\begin{table}[t]
\caption{Performance comparison between different methods on Caltech dataset (MR: Miss rate). }
%\small\addtolength{\tabcolsep}{-3pt}
\begin{center}
\begin{tabular}[c]{|c|c|}
\hline
\bf{Method} & MR(\%)  \\
\hline
JointDeep~\cite{ouyang2013joint} & 39.3 \\
\hline
SDN~\cite{luo2014switchable} & 37.9 \\
\hline
CifarNet~\cite{hosang2015taking} & 28.4 \\
\hline
LDCF~\cite{nam2014local} & 24.8 \\
\hline
AlexNet~\cite{hosang2015taking} & 23.3 \\
\hline
TA-CNN~\cite{tian2015pedestrian} &  20.9 \\
\hline
Checkerboards+~\cite{zhang2015filtered} & 17.1 \\
\hline
SA-FasterRCNN~\cite{li2018scale} & 9.7 \\
\hline
Proposed & 12.4 \\
\hline
\end{tabular}
\end{center}
\label{table:result2}
\end{table}

\begin{table}[t]
\caption{Performance comparison between different methods on INRIA dataset (MR: Miss rate). }
%\small\addtolength{\tabcolsep}{-3pt}
\begin{center}
\begin{tabular}[c]{|c|c|}
\hline
\bf{Method} & MR(\%)  \\
\hline
InformedHarr~\cite{zhang2014informed} & 14.43 \\
\hline
LDCF~\cite{nam2014local} & 13.79 \\
\hline
Franken~\cite{mathias2013handling} & 13.70 \\
\hline
Roerei~\cite{benenson2013seeking} & 13.53 \\
\hline
SA-FasterRCNN~\cite{li2018scale} & 8.04 \\
\hline
RPN+BF~\cite{zhang2016faster} & 6.88 \\
\hline
Proposed & 10.34 \\
\hline

\hline
\end{tabular}
\end{center}
\label{table:result3}
\end{table}

\begin{table}[t]
\caption{Performance comparison between different methods on ETH dataset (MR: Miss rate). }
%\small\addtolength{\tabcolsep}{-3pt}
\begin{center}
\begin{tabular}[c]{|c|c|}
\hline
\bf{Method} & MR(\%)  \\
\hline
JointDeep~\cite{ouyang2013joint} & 45 \\
\hline
LDCF~\cite{nam2014local} & 45 \\
\hline
Franken~\cite{mathias2013handling} & 40 \\
\hline
Roerei~\cite{benenson2013seeking} & 43 \\
\hline
TA-CNN~\cite{tian2015pedestrian} & 35 \\
\hline
RPN+BF~\cite{zhang2016faster} & 30 \\
\hline
Proposed & 31.12 \\
\hline
\end{tabular}
\end{center}
\label{table:result4}
\end{table}

\subsection{Comparisons with Other Deep Models}
\label{subsec:compare}
\textbf{Caltech:} We compare the performance of the proposed method with those of other deep models: JoinDeep~\cite{ouyang2013joint}, SDN~\cite{luo2014switchable}, LDCF~\cite{nam2014local}, TA-CNN~\cite{tian2015pedestrian}, Checkerboards+~\cite{zhang2015filtered}, and SA-FasterRCNN~\cite{li2018scale}. Table~\ref{table:result2} shows performance comparison between different methods on Caltech dataset. The proposed method performs the second by 12.4\% based on saliency and bounding box alignmen and achieves a slightly higher miss rate than SA-FasterRCNN~\cite{li2018scale}. \\

 \textbf{INRIA:} We also conduct performance comparison on INRIA dataset with InformedHaee~\cite{zhang2014informed}, LCDF~\cite{nam2014local}, Franken~\cite{mathias2013handling}, Roerei~\cite{benenson2013seeking}, and SA-FasterRCNN~\cite{li2018scale}. Table~\ref{table:result3} shows their performance on INRIA dataset. The INRIA dataset is a group of people-centric data rather than on real roads in a complex environment, which is much different from ETH or Caltech. It includes various types of data covering body parts, and is suitable for performance evaluation of body part detection and pedestrian detection from complex backgrounds. We evaluate the performance of the proposed method with part-level detection. As shown in Table.~\ref{table:result3}, the proposed method achieves comparable performance of 10.34\% to state-of-the-arts in a partially-occlusion dataset. \\

 \textbf{ETH:} ETH dataset is not a road environment, but it is worth assessing pedestrian detection performance by containing a large number of pedestrians. The proposed method shows a relatively low miss rate of 32.12\%. We compare our detector with JointDeep~\cite{ouyang2013joint}, LCDF~\cite{nam2014local}, Franken~\cite{mathias2013handling}, Roerei~\cite{benenson2013seeking}, TA-CNN~\cite{tian2015pedestrian} and RPN+BF~\cite{zhang2016faster}. Table~\ref{table:result4} shows performance comparison between them on ETH dataset. As shown in the table, the proposed method performs the second in MR (RPN+BF is the best) and achieves comparable performance to state-of-the-arts.

\section{Conclusions}
\label{sec:Conclusion}

In this paper, we have proposed part-level CNN for pedestrian detection using saliency and boundary box alignment. We have used saliency in the detection sub-network to remove false positives such as lamp posts and trees. We have utilized boundary box alignment in the alignment sub-network to recall the lost body parts. We have generated confidence maps using FCN and CAM, and estimated accurate position of pedestrians based on them. Experimental results demonstrate that the proposed method achieves competitive performance on Caltech, INRIA, and ETH datasets with state-of-the-art deep models for pedestrian detection in terms of MR.

In our future work, we will investigate pedestrian detection in low light condition such as night time with the help of near infrared (NIR) data.  

\bibliographystyle{IEEEtran}
\bibliography{IEEEabrv,refs}

% or


%\appendices

%\section{}
%Appendix two text goes here.\cite{Wolpert01thesupervised}

%\section*{Acknowledgment}

\ifCLASSOPTIONcaptionsoff
  \newpage
\fi

\end{document}